\documentclass{article}
\usepackage[margin=1.0in]{geometry}

\usepackage[utf8]{inputenc} 
\usepackage[T1]{fontenc}    
\usepackage{hyperref}       
\usepackage{url}            
\usepackage{booktabs}       
\usepackage{amsfonts}       
\usepackage{nicefrac}       
\usepackage{microtype}      

\usepackage{authblk}

\usepackage{amsmath}
\usepackage{dsfont} 
\usepackage{enumitem}
\usepackage{graphicx}
\usepackage{caption}
\usepackage{subcaption}
\usepackage{float} 
\usepackage{booktabs} 
\usepackage{multirow} 
\usepackage[table]{xcolor}
\usepackage{pdflscape} 
\usepackage{flushend}
\usepackage{algorithm}
\usepackage{algorithmic}
\usepackage{tikz}
\usetikzlibrary{bayesnet} 

\usepackage[round,numbers,sort&compress]{natbib}

\newcommand{\iid}{i.i.d. }
\newcommand{\ie}{i.e., }

\newcommand{\eg}{e.g., }
\newcommand{\Eg}{E.g., }

\newcommand{\eValue}[2]{\mathbb{E}_{#1}\left\{ #2 \right\}}

\newcommand{\N}[1]{\mathcal{N}\left( #1\right)}

\newcommand{\A}{\mathcal{A}}

\newcommand{\HH}{\mathcal{H}}

\newcommand{\argmax}{\mathop{\mathrm{argmax}}}
\newcommand{\argmin}{\mathop{\mathrm{argmin}}}

\title{Multi-armed bandits 
for resource efficient, online optimization of \\
language model pre-training: 
the use case of dynamic masking
\vspace*{1ex}
}

\author[1,2,*]{I\~nigo Urteaga}
\author[3]{Moulay-Za\"idane Dra\"idia}
\author[4,$\dagger$]{Tomer Lancewicki}
\author[5]{Shahram Khadivi}

\affil[1]{Basque Center for Applied Mathematics (BCAM), Bilbao, Spain}
\affil[2]{IKERBASQUE - Basque Foundation for Science, Bilbao, Spain}
\affil[*]{Work done while at the Department of Applied Physics and Applied Mathematics and the Data Science Institute, Columbia University, New York NY, USA}
\affil[3]{Data Science Institute, Columbia University, New York NY, USA}
\affil[4]{Walmart Global Tech,	USA}
\affil[$\dagger$]{Work done while at eBay Inc., San Jose (CA), USA}
\affil[5]{eBay Inc., Aachen, Germany}

\affil[ ]{ \newline
    {\tt \{iurteaga@bcamath.org, mad2314@columbia.edu, tomer.lancewicki@walmart.com, skhadivi@ebay.com\}} \newline 
    }

\begin{document}
\maketitle

\begin{abstract}
We design and evaluate a Bayesian optimization framework for resource efficient pre-training of Transformer-based language models (TLMs).
TLM pre-training requires high computational resources and introduces many unresolved design choices,
such as selecting its pre-training hyperparameters.
We propose a multi-armed bandit framework for the sequential selection of TLM pre-training hyperparameters,
aimed at optimizing language model performance, in a resource efficient manner.
We design a Thompson sampling algorithm,
with a surrogate Gaussian process reward model of the Masked Language Model (MLM) pre-training objective,
for its sequential minimization.
Instead of MLM pre-training with fixed masking probabilities,
the proposed Gaussian process-based Thompson sampling (GP-TS) accelerates pre-training
by sequentially selecting masking hyperparameters that improve performance.
We empirically demonstrate how GP-TS pre-trains language models efficiently,
\ie it achieves lower MLM loss in fewer epochs, across a variety of settings.
In addition, GP-TS pre-trained TLMs attain competitive downstream performance,
while avoiding expensive hyperparameter grid search.
GP-TS provides an interactive framework for efficient and optimized TLM pre-training that,
by circumventing costly hyperparameter selection,
enables substantial computational savings.
\end{abstract}

\section{Introduction}
\label{sec:submission}
In the field of Natural Language Processing (NLP),
models for learning unsupervised representations from unlabeled text based on Transformer architectures~\citep{vaswani2017attention}
are the state-of-the-art on a variety of tasks~\citep{kalyan2021ammus}.

Transformer-based language models (TLMs) like BERT~\citep{bert}, RoBERTa~\citep{roberta},
and their linage of advanced models~\citep{Amatriain2023},
rely on the combination of an unsupervised pre-training of the model, and a subsequent task-specific fine-tuning procedure. 
%
TLMs are pre-trained over large unlabeled text data using self-supervision, 
to learn the relationships between different sentences or words of the input.
Once the TLM is pre-trained over large volumes of data, it can be used in various downstream tasks, by fine-tuning task-specific model layers.
With pre-training, TLMs learn language representations
that are useful across downstream tasks,
minimizing the need and burden of retraining the entire model from scratch, again, for each task.
Extensive pre-training can lead to downstream performance improvements, \ie it is worth learning complex TLMs in huge natural language corpora before fine-tuning them for particular tasks. 

Many have replicated the pre-train-then-fine-tune strategy in different domains,
\eg pre-training BERT with scientific~\citep{j-Beltagy2019}
and biomedical corpora~\citep{j-Lee2020,j-Alsentzer2019,j-Gu2021};
or in-house, industry-specific TLMs~\citep{kalyan2021ammus}. 
In addition, continual pre-training
---taking a model pre-trained with general corpora to continue pre-training it with in-domain data---
is of great value,
yielding significant downstream gains~\citep{j-Gururangan2020}.

Even if conceptually simple and empirically powerful, pre-training is challenging and expensive.
Beyond the significant resources needed to pre-train the original BERT model by~\citet{bert},
the improvements of RoBERTa~\citep{roberta} relied on orders of magnitude higher computational resources~\citep{j-kaplan2020}.
In addition, the relationship between TLM architecture,
training corpus, pre-training hyperparameters, and evaluation metrics is complex and obscure. 
Therefore, previously overlooked pre-training design choices,
\eg pre-training hyperparameter selection,
result in significant performance differences.

With this work, we aim to improve the pre-training procedure of TLMs,
by \textit{sequentially} selecting hyperparameters that result in a more efficient and superior pre-training performance.
We hypothesize that an interactive selection of pre-training hyperparameters can accelerate and improve pre-training,
\ie we can achieve a better metric value in fewer epochs.
It is critical not only to achieve superior performance,
but to reduce the computational cost,
steering clear from time- and resource-expensive procedures.
Increased efficiency in TLM pre-training is paramount
amidst concerns pertaining to the carbon footprint of large language models~\citep{patterson2021carbon};
and specifically, the significant impact of hyperparameter selection
on resource utilization and power consumption~\citep{puvis-de-chavannes-etal-2021-hyperparameter}.

Our TLM pre-training use-case is \textit{random} dynamic masking of Masked Language Models (MLMs)
---in contrast to rule or task-based MLM dynamic masking solutions proposed in the literature~\citep{spanbert,ernie}.
Even though~\citet{roberta} showed the benefits of random dynamic masking,
the search for optimal masking hyperparameters is often carried out based on heuristic techniques and grid-based search.

In machine learning (ML), hyperparameter selection is commonly addressed as a black-box optimization problem,
which can be solved using
evolutionary algorithms~\citep{b-evolutionaryalgos},
entropy search methods~\citep{j-Hennig2012,ip-Hernandez-Lobato2014},
and Bayesian optimization (BO)~\citep{j-Frazier2018}.
In particular, BO can tackle the problem of optimizing an unknown objective function with possibly noisy evaluations~\citep{ip-Snoek2012},
and of speeding up resource allocation to promising hyperparameter configurations~\citep{hyperband}.
Aligned with the recent successes of~\citet{j-turner2021bayesian} in hyperparameter selection via Bayesian optimization,
we propose a BO framework for sequential tuning of MLM pre-training hyperparameters.
Our framework is different from BO techniques that speed up hyperparameter set evaluations,
such as Hyperband~\citep{hyperband}, which is a pure-exploration adaptive resource allocation algorithm
for apportioning resources among configurations in the non-stochastic setting.

We here cast the TLM pre-training procedure as a sequential decision process,
in which at each interaction, a reinforcement learning agent
selects an action (\eg pre-training hyperparameters) to maximize cumulative rewards (\eg the pre-training metric of interest).
To accommodate the black-box nature of the pre-training objective function,
we fit a probabilistic surrogate model to the empirical evaluations of the pre-training metric,
and propose a bandit-based technique for its sequential optimization.
In the MLM dynamic masking use case, the bandit actions are the dynamic masking probabilities;
and the MLM performance, the unknown function the bandit is trying to maximize,
based on estimates computed in the validation set.

Contrary to dynamic masking techniques that decide which subsets of tokens to mask via combinatorial optimization and dynamic programming~\citep{j-Vu2020};
we target online, sequential selection of masking hyperparameters for accelerated and improved pre-training.
In contrast to proposals that adapt the language model's masking policy to a particular task of interest~\citep{j-Kang2020},
we devise a generic online optimization framework that,
by sequential selection of MLM design choices,
provides fast and superior TLM pre-training performance, when pre-training ---from-scratch and continually--- across diverse corpora.

\paragraph*{The contributions}\hspace*{-2ex} of this work are:
\begin{itemize}[leftmargin=*]
	\item To present a bandit-based framework for efficient online optimization of TLM pre-training.
	Specifically, to formulate a Gaussian Process based Thompson sampling (GP-TS) algorithm for sequential MLM loss minimization.
	The novelty lays on modeling TLM pre-training validation losses with a Gaussian process reward model,
	and on formulating a Thompson sampling policy that minimizes them.
	
	\item To showcase empirically how GP-TS pre-trains TLMs better and faster:
	both when pre-training from-scratch and continually, across a variety of corpora.
	Besides, to show that GP-TS pre-trained TLMs provide top fine-tuned performance across diverse in-domain tasks, in fewer interactions.	
	
	\item To demonstrate that GP-TS's \textit{sequential selection} of how many tokens of the input to mask ---and how to mask them--- 
	results in improved and accelerated dynamic MLM pre-training, enabling significant resource utilization savings.	
\end{itemize}

To the best of our knowledge,
this work is the first
to address online optimization of TLM pre-training with bandit-based Bayesian optimization, 
and to showcase its performance and resource efficiency benefits.

The manuscript is organized as follows:
Section~\ref{sec:background} provides the background on Bayesian optimization, multi-armed bandits and TLM pre-training;
Section~\ref{sec:method} describes the proposed GP-TS method for TLM pre-training optimization;
with its empirical performance evaluated in Section~\ref{sec:experiments}.
Concluding remarks are provided in Section~\ref{sec:conclusion}.
\vspace*{-1ex}

\section{Background}
\label{sec:background}

\subsection{Bayesian optimization and bandits}
\label{ssec:mab}

\paragraph*{Bayesian optimization}\hspace*{-2ex} (BO) is a framework to address hyperparameter optimization in ML~\citep{ip-Snoek2012,ip-Klein2017,j-turner2021bayesian},
and many closely related applications~\citep{j-Negoescu2011,j-Calandra2016,ic-Frazier2016,ip-Hernandez-Lobato2017,j-Candelieri2018}.
BO relies on a probabilistic surrogate model of the objective function,
to tackle the problem of simultaneously fitting and optimizing a high-dimensional, non-convex function with unknown smoothness, and possibly
noisy evaluations~\citep{shahriari2015bayesian,j-Frazier2018}.
Due to the black-box nature of BO, the surrogate model must provide a measure of uncertainty, for which 
generative models, Bayesian neural networks and Gaussian processes are used~\citep{j-Maddox2021}.
Using this surrogate model, an acquisition function determines the next promising candidate to evaluate.
To address the challenge of learning about the environment (\ie exploration)
while simultaneously maximizing the observed outcomes (\ie exploitation),
the multi-armed bandit provides a useful framework~\citep{j-Lai1985}.

\paragraph*{The multi-armed bandit}\hspace*{-2ex} (MAB) is
an abstraction for problems that require learning while simultaneously maximizing attained rewards,
\ie balancing the exploration-exploitation tradeoff~\citep{b-Lattimore2020}.
A MAB is a sequential decision process
that requires decision-making under uncertainty~\citep{j-Slivkins2019}.

At each interaction $t=1,\cdots, T$,
a bandit agent chooses an action $a_t \in \A$ from a (not necessarily finite) set of actions $\A$,
and it observes stochastic reward $r_t$ drawn from an unknown distribution of the selected arm, $a_t$,
often characterized parametrically, $r_t\sim p(\cdot|a_t, \theta)$.
The MAB agent's goal is to maximize (expected) cumulative rewards, $R_T=\sum_{t=1}^T \mu_{a,t}$,
with each arm's expected reward denoted as $\mu_a = \eValue{p}{r|a,\theta}$.
The challenge is on the lack of knowledge about the reward generating mechanism,
which demands learning its properties (\eg its parameters), as it interacts with the environment.

A plethora of MAB algorithms have been proposed and analyzed over the years,
from computing optimal strategies~\citep{j-Gittins1979} and greedy approaches~\citep{j-Auer2002},
to upper confidence interval~\citep{j-Lai1987,ip-Kaufmann2012} 
and Thompson sampling~\citep{j-Thompson1935} algorithms.
%
For models in the exponential family,
the latter have been empirically and theoretically proven to perform competitively ~\citep{j-Lai1987,ip-Kaufmann2012,ip-Agrawal2012,ip-Agrawal2013,ic-Korda2013}, 
and extensions have been proposed
to model observed rewards 
via ensembles of models~\citep{ip-Lu2017},
Gaussian mixture models~\citep{ip-Urteaga2018, j-Urteaga2018},
Gaussian processes~\citep{ip-Srinivas2010,ip-Gruenewaelder2010},
and neural networks~\citep{ic-Osband2016}.

In the context of BO in general, and MABs in particular, reward uncertainty quantification is critical.
Gaussian processes~\citep{b-Rasmussen2005} provide not only adequate Bayesian uncertainty estimates,
but a flexible solution for surrogate models that encode smoothness assumptions of the payoff function~\citep{ip-Krause2011, ip-Bogunovic2016, ip-Nguyen2020}.
We resort to a Gaussian process reward model in the proposed bandit-based BO framework for TLM pre-training.

\subsection{Language model pre-training and the Masked Language Model}
\label{ssec:roberta_pretraining}

Pre-training enables learning representations that generalize across tasks,
\ie it allows for a language model to be better initialized for quick fine-tuning (while avoiding overfitting) to downstream tasks.
TLMs learn language representations in pre-training based on one (or more) self-supervised task.
Two popular pre-training objectives are Masked Language Model (MLM) and Next Sentence Prediction (NSP)~\citep{bert}.
We focus on MLM pre-training as in~\citep{bert,roberta};
where for an input sequence of words or tokens,
a random sample of the tokens is replaced with the $[MASK]$ token,
and the goal is to predict them.

For an input sequence $d$ of $N$ tokens, with special tokens delimiting them,
\begin{equation}
d \equiv [CLS], q_1, \cdots , q_N, [EOS]
\end{equation}
MLMs select a random sample of the tokens $q_{i}, i=\{1, \cdots, N\}$, replace them with the mask, 
and learn to predict these masked tokens.
For pre-training the original BERT model~\citep{bert}, a random but \textit{static} subset of the input sequence tokens was replaced with the mask.

~\citet{roberta} proposed a \textit{dynamic} masking procedure,
which generates a new masking pattern (given a fixed probability of masking) for every input sequence.
\citet{roberta} demonstrate that this dynamic approach is beneficial when pre-training for more steps or with larger datasets.

Dynamic masking relies on several hyperparameters:
($i$) the probability $\rho$ of replacing an input token with the mask,
($ii$) the probability $\gamma$ that a masked token is left unmasked,
and ($iii$) the probability $\lambda$ of replacing a token with a random token, instead of with the mask.
Online optimization of these hyperparameters $\psi=\left(\rho, \gamma, \lambda\right)$ is the use-case for our experiments in Section~\ref{sec:experiments}.

\paragraph*{MLM pre-training}\hspace*{-2ex}
aims at minimizing the MLM loss:
a function of the original ($D$) and masked ($\widehat{D}$) datasets,
the TLM architecture with its parameters $w\in W$,
and pre-training hyperparameters $\psi\in\Psi$.

The MLM objective is the cross-entropy loss of predicting the masked tokens in the masked sequence $\widehat{d}\in\widehat{D}$, where we denote with $m_{i}=\{0,1\}$ whether tokens $q_i, i=\{1, \cdots, N\}$, from the original input sequence $d \in D$ have been masked in $\widehat{d}$:
\begin{align}
l(d, \widehat{d}; w, \psi) &= -\log p(d|\widehat{d}; w, \psi)
= -\sum_{i=1}^N m_{i} \log p(q_i|\widehat{q_i}; w, \psi) = -\hspace*{-1ex} \sum_{i=1}^N m_{i} \log \hspace*{-0.5ex}\left( \frac{e^{\left(\chi(\widehat{q_i};w, \psi)^\top \xi(q_i)\right)}}{\sum_{i^\prime=1}^{N} e^{\left(\chi(\widehat{q_{i}^\prime};w, \psi)^\top \xi(q_{i}^\prime)\right)}}\hspace*{-0.5ex}
\right) \;,
\label{eq:mlm_loss}
\end{align}
$\chi(\widehat{q_{i}};w, \psi)$ denotes the TLM's representation of the masked token $q_i$,
and $\xi(q_i)$ is its original embedding.

The pre-training objective is to find the TLM that minimizes the MLM loss between the original dataset $D$ and its masked version $\widehat{D}$.
%
In practice, this minimization is executed via stochastic gradient-descent,
run for $e=1,\cdots, E,$ epochs with random mini-batches $D_{e} \in D$ per epoch $e$,
$
\widehat{w_e}=\argmin_{w \in W} l(D_{e}, \widehat{D_e}; w, \psi) \;.
$

The analytical form of the MLM loss, a function of selected hyperparameters $\psi$ and the data where it is evaluated, is in general complex and unknown.
However, estimates of the MLM loss are available at every pre-training epoch $e$.
Namely, an empirical estimate of the MLM loss can be computed in the validation set.
For fair comparisons under different training setups (\eg mini-batch sizes and hyperparameters), per-epoch \textit{averaged} empirical MLM losses are computed in the validation dataset $D_{val}$,
\begin{align}
&\bar{l}(D_{val};\psi)=\bar{l}(D_{val}, \widehat{D_{val}}; w, \psi) = - \sum_{d \in D_{val}} \frac{\sum_{i=1}^{N_d} m_{i} \log p(q_i|\widehat{q_i}; w, \psi)}{\sum_{i^\prime=1}^{N_d} m_{i^\prime}} \; ,
\label{eq:mlm_averagedloss}
\end{align}
where we drop the dependency with respect to TLM parameters $w$ and the masked validation dataset $\widehat{D_{val}}$ to avoid notation clutter.

\section{Proposed bandit-based framework}
\label{sec:method}
We cast TLM pre-training as a sequential decision process,
to be solved by a multi-armed bandit agent
that interactively optimizes the analytically unknown pre-training loss, 
based on its sequentially observed empirical evaluations.
We define pre-training steps,
\ie a fixed number of stochastic gradient updates $u$ in the training set,
as bandit interactions $t=1,\cdots,T$.
The goal is to minimize the TLM pre-training objective $l(\cdot |\psi)$ given tunable hyperparameters $\psi$,
with (stochastic) evaluations of the loss function in the validation set.

Pre-training hyperparameters at interaction $t$, $\psi_t$, are the bandit's arms, \ie $a_t=\psi_t$.
%
For MLM pre-training with dynamic masking,
at each bandit interaction,
the agent selects hyperparameters $\psi$ (the proportion of tokens to mask and their masking probabilities),
pre-trains the TLM for certain stochastic updates to minimize the MLM loss, 
and evaluates its performance in the validation subset,
as per Equation~\eqref{eq:mlm_averagedloss}.
To accommodate the black-box nature of the pre-training objective, for which only stochastic evaluations are available,
we formulate a surrogate reward function (leveraging empirical MLM validation loss estimates)
for the bandit to maximize, as it sequentially selects which arm to play.

\subsection{From MLM pre-training to Gaussian process-based regret minimization}
\label{ssec:method_rewards}
We transform the empirical pre-training validation loss at each MAB interaction
into a reward quantity for it's sequential minimization by the bandit agent.
Specifically, we compute bandit rewards as the normalized difference in averaged empirical MLM losses between bandit interactions, \ie

\begin{align}
r_t (\psi_t) &= \frac{
	[- \bar{l}_t(D_{val}; \psi_t)] 
		- [- \bar{l}_{t-1}(D_{val}; \psi_{t-1})]
	}{
		[- \bar{l}_{t-1}(D_{val}; \psi_{t-1})]
	} \;.
\label{eq:reward_mlm_delta}
\end{align}

By normalizing reward differences per-interaction,
we mitigate the potential non-stationary effect sequentially selected hyperparameters might have on TLM pre-training.
With rewards as (normalized) empirical MLM loss differences,
we capture how much (relative) improvement each action provides.

Rewards in Equation~\eqref{eq:reward_mlm_delta}
are based on stochastic draws from an analytically unknown objective function,
\ie only empirical estimates $\bar{l}_t(\cdot)$ of the MLM objective are available.
%
To accommodate these noisy observations of the unknown loss function $l(\cdot|\psi)$
---that we aim at optimizing with respect to its hyperparameters $\psi$---
we model the bandit reward function via a Gaussian process (GP) model $f(\cdot ;\theta)$ of the pre-training objective,
with observed rewards independent and identically (i.i.d.) distributed as
\vspace*{-1ex}
\begin{align}
r_t(\psi_t) &=f(\psi_t ; \theta) + \epsilon_t \;,
\label{eq:rewards_gp}
\vspace*{-2ex}
\end{align} 
where $\epsilon_t$ denotes the stochastic nature of each of the observed rewards ---based on empirical estimates computed in Equation~\eqref{eq:reward_mlm_delta}.
Hence, we overcome the black-box nature of the pre-training objective (\eg the MLM loss) by modeling observed rewards as realizations of a noisy surrogate GP model~\citep{b-Rasmussen2005}.

The mean $\mu(\cdot)$ and kernel functions $k(\cdot,\cdot)$ of a GP $f(\cdot) \sim GP(\mu(\cdot), k(\cdot,\cdot))$ determine the reward function class:
\ie the regularity and smoothness of the pre-training loss.
These are parameterized prior-functions $\mu(\cdot|\theta_{\mu})$ and $k(\cdot, \cdot|\theta_k)$,
which can be fitted to the observed data $r_{1:T} = (r_1, \cdots, r_T)$ at inputs $\psi_{1:T} = (\psi_1, \cdots, \psi_T)$~\citep{b-Rasmussen2005}.
For instance, via Type-II maximum likelihood estimation (MLE) of the GP parameters $\theta=(\theta_{\mu}, \theta_k)$,
$
\hat{\theta} =\argmax_{\theta} \log p\left(r_{1:T}|f(\psi_{1:T} | \theta) \right) 
$,
where the data likelihood $p(r|f (\cdot; \theta))$ is a function of the observation noise probability distribution.
Given a fitted GP, posterior inference
---computing the predictive distribution of a new datapoint $\psi^\prime$ after observing $\psi_{1:T}$---
can be performed in closed or approximate form~\cite{ip-Titsias2009,ip-Flaxman2015,ip-Pleiss2018}.

\subsection{GP-Thompson sampling for TLM pre-training.}
\label{ssec:method_gpts}
Leveraging the GP reward model in Equation~\eqref{eq:rewards_gp},
we devise a bandit-based interactive method that executes a Thompson sampling (TS) policy
for TLM pre-training optimization.
We resort to Thompson sampling~\cite{j-Russo2018} due to both its implementation flexibility and efficiency,
as well as its competitive empirical performance with theoretical guarantees in many settings~\cite{ip-Agrawal2013,ip-Krause2011,ip-Nguyen2020,ip-Srinivas2010}.

The proposed Gaussian process-based Thompson sampling (GP-TS)
---with pseudo-code provided in Algorithm~\ref{alg:ts_pretrain_hyperparams}---
views the TLM pre-training objective as an unknown black-box function with inputs $a_t=\psi_t$ and outputs $r_t(\psi_t)$ as in Equation~\eqref{eq:reward_mlm_delta}.
GP-TS makes decisions on what bandit arm $a_t=\psi_t$ to play at each TLM pre-training interaction $t=1,\cdots,T,$
informed by its GP reward model of Equation~\eqref{eq:rewards_gp}, 
to maximize its observed cumulative rewards $R_T=\sum_{t=1}^T r_{t}(\psi_t)$.

\begin{algorithm}
	\caption{GP-TS for TLM pre-training}
	\label{alg:ts_pretrain_hyperparams}
	\begin{algorithmic}[1]
		\STATE {\bfseries Input}: TLM and pre-training corpus
		\STATE {\bfseries Input}: Pre-training hyperparameter space $\Psi$
		\STATE {\bfseries Input}: Number of pre-training interactions $T$, number of updates per-interaction $u$
		\STATE {\bfseries Input}: GP prior functions $\mu(\cdot)$ and $k(\cdot, \cdot)$, \\ with initial hyperparameters $\theta_0$		
		\STATE {\bfseries Initialize}: $\A=\Psi$, $\hat{\theta}_1=\theta_0$, $\HH_1=\emptyset$
		\FOR{$t=1, \cdots, T$}
		\STATE Draw posterior sample from GP, 
			$\mu_{a}^{(t)} \sim f(\mu_t(a|\hat{\theta}_t), k_t(a, a^\prime|\hat{\theta}_t)) \;.$
		\STATE Select arm based on drawn posterior sample, 
			$a_{t}=\argmax_{a^\prime \in \A} \mu_{a^\prime}^{(t)} \;.$
		\STATE Run TLM pre-training for $u$ steps, with hyperparameters $\psi_t=a_t \;.$
		\STATE Compute pre-trained TLM validation loss, 
			$\bar{l}_t(D_{val};\psi_t)$ as in Equation~\eqref{eq:mlm_averagedloss}.
		\STATE Observe bandit reward, 
			$r_{t}(\psi_t)$ as in Equation~\eqref{eq:reward_mlm_delta}.
		\STATE Update bandit history 
			$\HH_{1:t}=\HH_{1:t-1} \cup \left\{a_{t}=\psi_t, r_{t}(\psi_t)\right\} \;.$
		\STATE Fit GP model with $\HH_{1:t}$, 
			$\hat{\theta}_{t+1} =\argmax_{\theta} \log p\left(r_{1:t}|f(\psi_{1:t} ; \theta) \right) \;.$
		\ENDFOR
	\end{algorithmic}
\end{algorithm}

GP-TS accommodates continuous arms $a_t=\psi_t$,
with dimensionality determined by the pre-training hyperparameter space $\psi \in \Psi$.
Any TLM can be used within the proposed framework,
as long as the hyperparameter space $\psi \in \Psi$ is identified,
and rewards as in Equation~\eqref{eq:reward_mlm_delta} are computed for a pre-training objective $l(\cdot|\psi)$ of interest.

GP-TS draws predictive function samples for the next TLM pre-training interaction
from its GP reward model posterior,
updated at every bandit interaction as indicated in Step 7 of Algorithm~\ref{alg:ts_pretrain_hyperparams}.
As in other TS methods, these samples are used to determine ---in Step 8 of Algorithm~\ref{alg:ts_pretrain_hyperparams}---
the arms (hyperparameters $\psi_t$) to be used in the next bandit interaction.
After $u$ pre-training steps\footnote{
	Note that $u$ stochastic gradient updates might or might not correspond to a full pre-training epoch $e$.
},
the model's MLM validation loss is computed
to evaluate the observed bandit rewards $r_{t}(\psi_t)$ of Equation~\eqref{eq:reward_mlm_delta}.
After each interaction $t$, new evidence is collected in Step 12
to re-fit the GP model to the observed input (action)-output (rewards) history $\HH_{1:t}$.
For instance, via Type-II MLE as in Step 13 of Algorithm~\ref{alg:ts_pretrain_hyperparams},
although other GP parameter optimization procedures might be used
---see Appendix~\ref{asec:GP_details} for details on GP models and posterior inference.

\section{Experiments}
\label{sec:experiments}

\subsection{Evaluation set-up}
\label{ssec:set_up}

We probe the ability of the proposed GP-TS to,
given a dataset, a TLM architecture, and a computational budget,
efficiently pre-train well-performing language models.
We scrutinize pre-training performance of a specific TLM architecture under equal experimental conditions 
and do not compare performance to state-of-the-art, large-scale TLMs.

For our experiments,
we incorporate RoBERTa~\citep{roberta} as implemented by~\citet{fairseq}
in our Python implementation of GP-TS\footnote{
	Code available at~\href{https://github.com/iurteaga/gp_ts_nlp}{https://github.com/iurteaga/gp\_ts\_nlp}.
} as in Algorithm~\ref{alg:ts_pretrain_hyperparams}
---Appendix~\ref{asec:implementation_details_gp} provides implementation and configuration details.
We compare pre-training performance of RoBERTa models
based on a grid-search over masking hyperparameters ---as executed by~\citet{roberta}---
to RoBERTa models pre-trained by GP-TS\footnote{
	We do not execute any other hyperparameter optimization.
}.
We focus our evaluation on MLM validation loss and downstream per-task accuracy metrics,
and report the negligible computational overhead of pre-training with GP-TS in Appendix~\ref{asec:computational_overhead}. 

We study two variants of GP-TS, depending on the masking hyperparameters it optimizes:
\begin{enumerate}
	\vspace*{-1ex}
	\item \texttt{GP-TS $\rho$}, where the bandit arm is the masking probability $\rho$ of replacing an input token with the \textit{mask} token
	(other hyperparameters are fixed to default $\gamma=0.1$ and $\lambda=0.1$ values as in~\citet{roberta});
	and 
	\vspace*{-1ex}
	\item \texttt{GP-TS $\psi=\left(\rho, \gamma, \lambda\right)$},
	where GP-TS optimizes over all MLM dynamic masking hyperparameters:
	the bandit search space is a three-dimensional hypercube $\Psi$
	with no expert guidance. 
	\vspace*{-1ex}
\end{enumerate}

\paragraph*{Pre-training datasets.} \hspace*{-2ex}
We gather three distinct datasets, two based on publicly available corpora,
and one based on private data from eBay:

\begin{itemize}[leftmargin=*]
	\item \textbf{wiki-c4}: We pre-process and encode publicly available Wikitext-103~\citep{wikitext103} 
	and 
	Google's c4 RealNews~\citep{c4_realnews} datasets
	for pre-training, from scratch, each of TLM.
	This corpora is similar to those originally used by~\citet{bert} and \citet{roberta}.
	
	\item \textbf{mimic}: We pre-process and encode free-text clinical notes available in the public MIMIC-III Clinical database~\citep{mimic}, which contains deidentified nursing and physician notes, ECG and imaging reports, and discharge summaries for patients who stayed in intensive care units at Beth Israel Deaconess Medical Center.
	
	\item \textbf{e-commerce}: We pre-process and encode a random subset of eBay marketplace inventories, which contains different product titles and descriptions provided by marketplace users, as well as category tags associated with each item and product reviews.

\end{itemize}

Each dataset contains text of very different linguistic characteristics and sizes (see summary statistics in Appendix~\ref{asec:pretraining_dataset_details}),
which we leverage to investigate TLM pre-training across a variety of settings.

We evaluate candidate TLMs
($i$) when pre-training \textit{from-scratch}, \ie from a randomly initialized architecture; and
($ii$) with \textit{continual} pre-training, \ie when continuing pre-training a TLM architecture previously trained in other NLP corpora~\citep{kalyan2021ammus}.
Continual pre-training results we present are for the RoBERTa-base architecture as pre-trained~\citet{robertabase_fairseq}
that we continue to pre-train in our domain-specific datasets, \ie \texttt{mimic} and \texttt{e-commerce}.

\vspace*{-1ex}
\paragraph*{Fine-tuning in downstream tasks.} \hspace*{-2ex}
Pre-trained language models are most useful when applied to downstream tasks,
as there is no need to retrain the entire model again.
We evaluate pre-trained TLM's in the following in-domain tasks\footnote{
	We abstain from fine-tuning RoBERTa-base models, pre-trained with \texttt{wiki-c4} data only, in downstream Glue tasks~\citep{glue},
	as these would not match state-of-the-art results due to both the size-limited pre-training dataset, and the model architecture used.
}:

\begin{itemize}[leftmargin=*]
	\item \textbf{e-commerce title classification}: A binary classification task to decide whether a pair of item titles belong to the same marketplace product.
	Item titles are instances of a product sold by a specific seller, which can have different attributes like condition or can exist as a special version (\eg a signed book), yet refer to the same product.
	
	\item \textbf{e-commerce title similarity}: A task using the same title-pair data as above, but formulated as a similarity task.
	Namely, we learn a distance metric between item titles to help discriminate whether they belong or not to the same product.
	
	\item \textbf{e-commerce title quality}: A classification task that predicts if a title fulfills the marketplace requirements for it to be a product title.
	Titles must contain the product's main relevant information
	---the brand, the product name and/or type, and all distinguishable attributes, \ie its key features---
	but should not contain conditions, marketing terms, or any other non-product related information.
	
	\item \textbf{medical MLI}: A natural language inference task annotated by doctors~\citep{medMLI},
	which is grounded in the medical history of patients collected in MIMIC-III~\citep{mimic}.
	It contains sentence pairs ---the premise and the hypothesis statements--- with a corresponding label indicating their inferential relationship (\eg entailment, contradiction, or neutral).
\end{itemize}

Summary statistics for each in-domain per-task dataset
are provided in Appendix~\ref{asec:finetuning_dataset_details}.

To elucidate how the pre-trained TLMs' quality evolves over pre-training interactions,
we fine-tune (for ten epochs) the pre-trained RoBERTa models at each pre-training interaction $t$.
We report the best classification accuracy of each fine-tuned model across pre-training interactions and fine-tuning epochs.

\subsection{GP-TS pre-training of RoBERTa models}
\label{ssec:pretraining}

\begin{figure*}[!th]
	\centering
	\begin{subfigure}[c]{0.32\textwidth}
		\includegraphics[width=\textwidth]{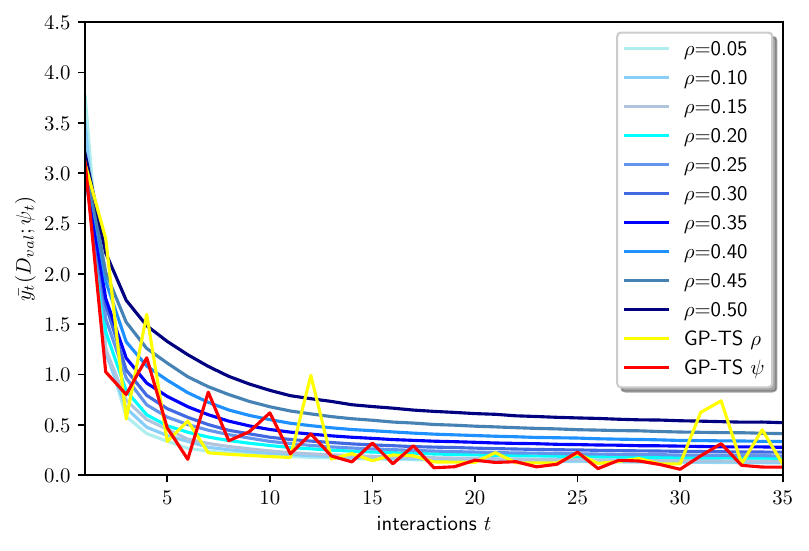}
		\vspace*{-4ex}
		\caption{\texttt{wiki-c4}.}
		\label{fig:pretraining_new_wikic4}
	\end{subfigure}
	\begin{subfigure}[c]{0.32\textwidth}
		\includegraphics[width=\textwidth]{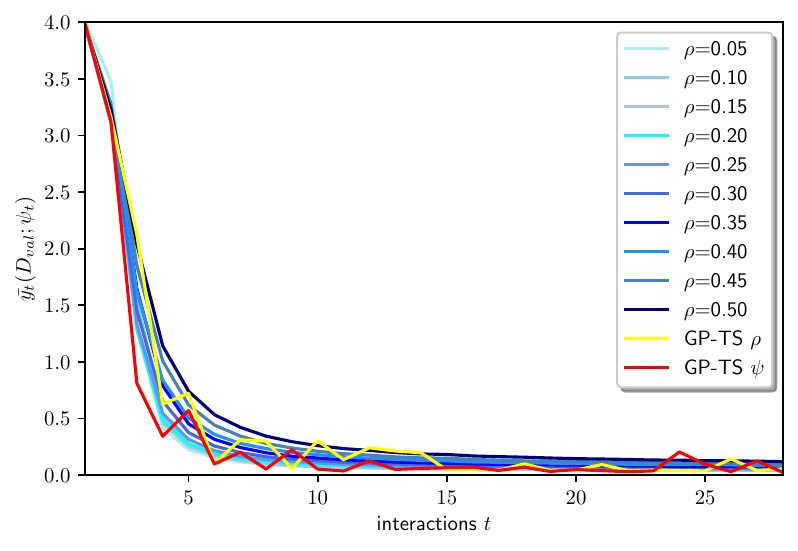}
		\vspace*{-4ex}
		\caption{\texttt{mimic}.}
		\label{fig:pretraining_new_mimic}
	\end{subfigure}
	\begin{subfigure}[c]{0.32\textwidth}
		\includegraphics[width=\textwidth]{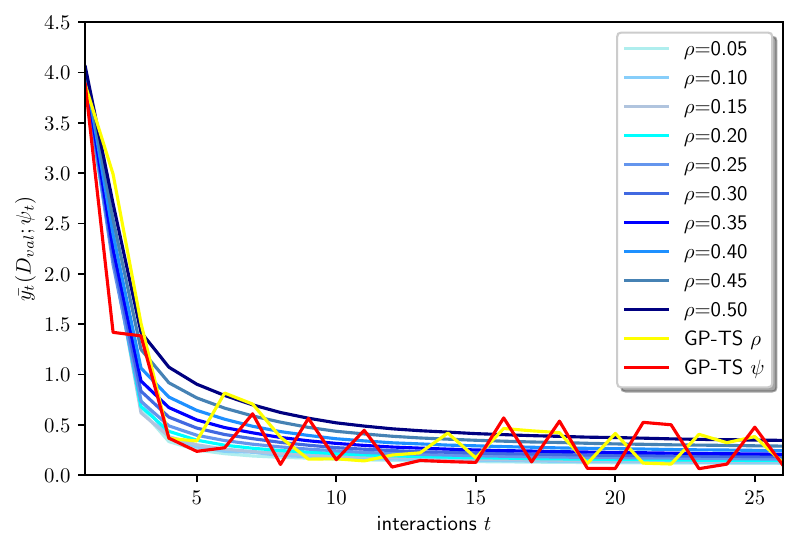}
		\vspace*{-4ex}
		\caption{\texttt{e-commerce}.}
		\label{fig:pretraining_new_ebay}
	\end{subfigure}
	\vspace*{-1ex}
	\caption{MLM validation loss comparison (lower is better) of grid-search and GP-TS based \textit{from-scratch} pre-trained RoBERTa models, over interactions.
	}
	\label{fig:new_pretraining}
\end{figure*}

We compare \textit{from-scratch} pre-training performance of all RoBERTa models
---pre-trained with fixed hyperparameters or by GP-TS--- in Figure~\ref{fig:new_pretraining},
where we illustrate MLM validation losses of each model over pre-training interactions:
GP-TS attains the lowest MLM loss values in fewer interactions.
Recall that when pre-training TLMs, validation performance varies across training epochs;
hence, we are interested in identifying the best pre-trained model
---as per the lowest validation metric---
instead of selecting the pre-trained TLM available at the last training epoch.

\begin{figure}[!h]
	\centering
	\begin{subfigure}[c]{0.32\textwidth}
		\includegraphics[width=\textwidth]{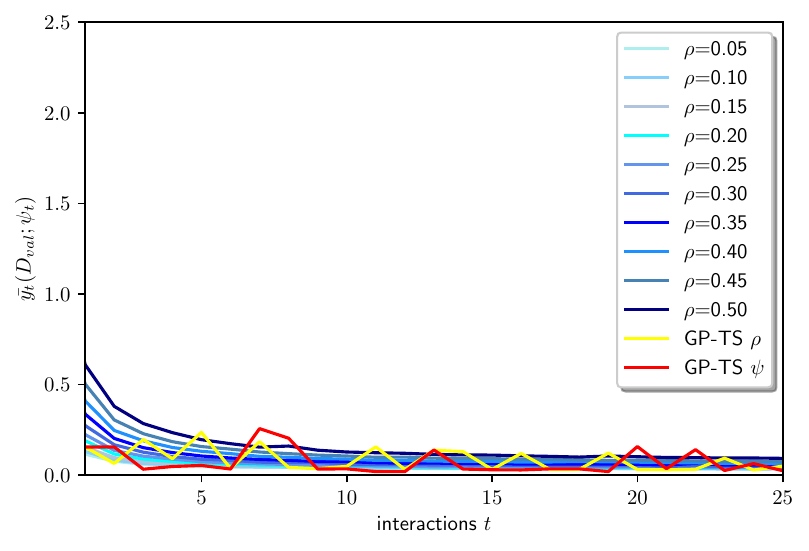}
		\vspace*{-4ex}
		\caption{\texttt{mimic}.}
		\label{fig:continual_pretraining_mimic}
	\end{subfigure}
	\begin{subfigure}[c]{0.32\textwidth}
		\includegraphics[width=\textwidth]{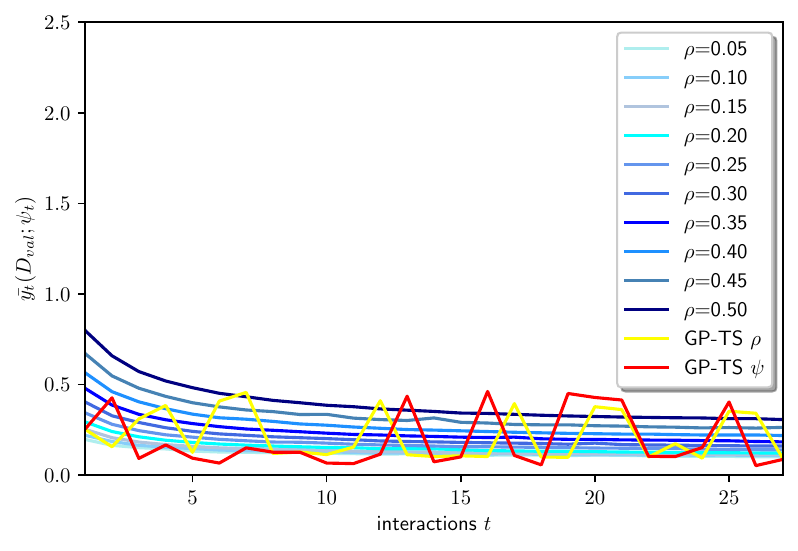}
		\vspace*{-4ex}
		\caption{\texttt{e-commerce}.}
		\label{fig:continual_pretraining_ebay}
	\end{subfigure}
	\vspace*{-1ex}
	\caption{MLM validation loss comparison (lower is better) of grid-search and GP-TS based \textit{continually} pre-trained RoBERTa models over interactions.
	}
	\label{fig:continual_pretraining}
\end{figure}

Results for \textit{continual} pre-training are provided in Figure~\ref{fig:continual_pretraining},
where we observe that GP-TS continually pre-trains the best performing RoBERTa models ---the fastest--- for both in-domain datasets.

MLM validation losses for models pre-trained with GP-TS fluctuate across interactions,
depending on the stochastic action (hyperparameter value) selected by the GP-TS agent.

We evaluate the influence of different realizations of GP-TS (with different random seeds) in Table~\ref{tab:medical_MLM_medical_robertabase},
where we observe that GP-TS always pre-trains models with the lowest MLM loss, and in less interactions (indicated within parentheses).
Practitioners are interested in using the model with the lowest validation MLM loss,
which GP-TS consistently finds across all studied datasets and pre-training approaches,
in fewer pre-training interactions.

\begin{table}[!h]
	\caption{Best MLM loss attained before interactions 20 and 30, when pre-training RoBERTa models continually in the medical domain corpora.}
	\label{tab:medical_MLM_medical_robertabase}
	\vspace*{-1ex}
	\centering
		\begin{tabular}{|c|c|c|}
			\hline
		\cellcolor[gray]{0.6} 
			& By interaction 20 \cellcolor[gray]{0.6} 
			& By interaction 30 \cellcolor[gray]{0.6} \\ \cline{1-3}
		\cellcolor[gray]{0.6} 
			& Best MLM loss\cellcolor[gray]{0.8} 
			& Best MLM loss \cellcolor[gray]{0.8} \\ \cline{1-3}
		\multirow{-3}{*}{Model}\cellcolor[gray]{0.6} 
			& (at interaction)\cellcolor[gray]{0.8} 
			& (at interaction)\cellcolor[gray]{0.8} \\ \hline
$\rho$=0.05 	 & 0.04 (18) 	 & 0.037 (28) 	\\ \hline 
$\rho$=0.10 	 & 0.04 (18) 	 & 0.036 (27) 	\\ \hline 
$\rho$=0.15 	 & 0.044 (18) 	 & 0.038 (27) 	\\ \hline 
$\rho$=0.20 	 & 0.048 (18) 	 & 0.042 (28) 	\\ \hline 
$\rho$=0.25 	 & 0.054 (19) 	 & 0.046 (27) 	\\ \hline 
$\rho$=0.30 	 & 0.066 (18) 	 & 0.056 (27) 	\\ \hline 
$\rho$=0.35 	 & 0.076 (19) 	 & 0.064 (29) 	\\ \hline 
$\rho$=0.40 	 & 0.091 (19) 	 & 0.077 (29) 	\\ \hline 
$\rho$=0.45 	 & 0.113 (19) 	 & 0.095 (29) 	\\ \hline 
$\rho$=0.50 	 & 0.134 (19) 	 & 0.112 (27) 	\\ \hline 
\rowcolor[gray]{0.95} 
GP-TS $\rho$ (seed 1) 	 & 0.037 (14) 	 & 0.033 (20) 	\\ \hline 
\rowcolor[gray]{0.95}
GP-TS $\rho$ (seed 2) 	 & 0.036 (19) 	 & 0.033 (28) 	\\ \hline 
\rowcolor[gray]{0.95}
GP-TS $\rho$ (seed 3) 	 & 0.038 (14) 	 & 0.032 (21) 	\\ \hline 
\rowcolor[gray]{0.95}
GP-TS $\rho$ (seed 4) 	 & 0.032 (18) 	 & 0.032 (18) 	\\ \hline 
\rowcolor[gray]{0.95}
GP-TS $\rho$ (seed 5) 	 & 0.038 (13) 	 & 0.032 (20) 	\\ \hline 
GP-TS $\psi$ (seed 1) 	 & 0.027 (8) 	 & 0.019 (21) 	\\ \hline 
GP-TS $\psi$ (seed 2)	 & \textbf{0.02 (15)} 	 & 0.02 (15) 	\\ \hline 
GP-TS $\psi$ (seed 3)	 & \textbf{0.02 (17)} 	 & 0.019 (28) 	\\ \hline 
GP-TS $\psi$ (seed 4)	 & 0.036 (14) 	 & 0.019 (21) 	\\ \hline 
GP-TS $\psi$ (seed 5)	 & \textbf{0.02 (16)} 	 & \textbf{0.018 (28)} 	\\ \hline 
		\end{tabular}
\vspace*{-1ex}
\end{table}

GP-TS not only circumvents the need for costly grid searches, but enables improved performance:
it attains reduced MLM loss at earlier interactions than grid-search baselines.
Recall how GP-TS $\psi$ outperforms all the alternatives in Table~\ref{tab:medical_MLM_medical_robertabase},
as it pre-trains models with the lowest MLM, the fastest
---even when no good initial guesses for the MLM hyperparameters $\psi=\left(\rho, \gamma, \lambda\right)$ are available.

In summary, the benefits of interactive GP-TS pre-training do not pertain to the attained MLM values only,
but to an accelerated, efficient procedure.
We emphasize the computational efficiency of GP-TS:
it adds little to no overhead ---details on the computational cost of GP-TS are provided in Appendix~\ref{asec:computational_overhead}---
while providing clear benefits for language model pre-training.
It attains best MLM pre-training performance in less interactions,
avoiding computationally expensive hyperparameter search.

To the best of our knowledge, these experiments provide novel evidence that,
instead of MLM pre-training with fixed masking hyperparameters,
sequentially deciding which masking values to use is beneficial.
Namely, GP-TS finds \textit{sequences} of dynamic masking hyperparameters
(when optimizing over $\rho$ or a three-dimensional hyperparameter space $\psi\in\Psi$)
that minimize MLM loss across datasets, when pre-training from-scratch and continually.


\subsection{GP-TS pre-trained RoBERTa models for downstream fine-tuned tasks}
\label{ssec:downstream}

We scrutinize how performant in-domain GP-TS pre-trained RoBERTa models are,
when compared to grid-search based models,
after in-domain per-task fine-tuning.
We note that the downstream, fine-tuned performance of RoBERTa models pre-trained from-scratch with in-domain data is, as expected, lower than if continually pre-trained.

The fine-tuned accuracy of continually pre-trained models
of Figure~\ref{fig:continual_pretraining} are presented in Table~\ref{tab:ebay_tasks_robertabase}:
we showcase best (per-task) test-set accuracy for each fine-tuned model,
and at which pre-training interaction was such value attained.
Results are computed on each per-task test-set,
\ie a subset of each task's dataset (see details in Table~\ref{tab:finetuning_dataset_details})
that has not been used for fine-tuning nor hyperparameter optimization.

\begin{table}[!h]
	\caption{Best fine-tuned, downstream task test-set accuracy (higher is better) for continually pre-trained RoBERTa models.
		The first row corresponds to the fine-tuned performance of the RoBERTa model from which continual pre-training is started.}
	\label{tab:ebay_tasks_robertabase}
	\centering
		\begin{tabular}{|c|c|c|c|c|}
			\hline
			\cellcolor[gray]{0.6} 
			& e-commerce \cellcolor[gray]{0.6} 
			& e-commerce \cellcolor[gray]{0.6} 
			& e-commerce \cellcolor[gray]{0.6}
			& medical \cellcolor[gray]{0.6} \\ \cline{1-5}
			\cellcolor[gray]{0.6} 
			& title classification \cellcolor[gray]{0.6} 
			& title similarity \cellcolor[gray]{0.6} 
			& title quality \cellcolor[gray]{0.6}
			& MLI \cellcolor[gray]{0.6} \\ \cline{1-5}
			\cellcolor[gray]{0.6} 
			& Accuracy\cellcolor[gray]{0.8} 
			& Accuracy\cellcolor[gray]{0.8} 
			& Accuracy\cellcolor[gray]{0.8} 
			& Accuracy\cellcolor[gray]{0.8} \\ \cline{1-4}
			\multirow{-4}{*}{Model}\cellcolor[gray]{0.6} 
			& (at interaction)\cellcolor[gray]{0.8} 
			& (at interaction)\cellcolor[gray]{0.8} 
			& (at interaction)\cellcolor[gray]{0.8} 
			& (at interaction)\cellcolor[gray]{0.8} \\ \hline
RoBERTa base 	 & 97.2 \;\;(0) 	 & 97.2 \;\;(0) 	 & 75.1 \;\;(0) 	 & 67.5 \;\;(0) 	\\ \hline 
$\rho$=0.05 	 & 97.8 (26) 	 & 97.8 (26) 	 & 77.6 (15) 	 & 72.9 \;\;(3) 	\\ \hline 
$\rho$=0.10 	 & 97.9 (27) 	 & 97.9 (27) 	 & 77.7 (15) 	 & 71.9 \;\;(9) 	\\ \hline 
$\rho$=0.15 	 & 97.8 (13) 	 & 97.8 (13) 	 & 77.7 (18) 	 & 72.5 (13) 	\\ \hline 
$\rho$=0.20 	 & 97.8 \;\;(8)	 & 97.8 \;\;(8) 	& 77.4 (10) 	 & 73.3 (14) 	\\ \hline 
$\rho$=0.25 	 & 97.9 (17) 	& 97.9 (17) 	 & 77.7 \;\;(6)	 & 72.9 (12) 	\\ \hline 
$\rho$=0.30 	 & 97.9 (19)	& 97.9 (19) 	& 77.8 \;\;(7)	 & \textbf{73.2 \;\;(7)} 	\\ \hline 
$\rho$=0.35 	 & 97.9 \;\;(9)	 & 97.9 \;\;(9) 	& 77.8 (18) 	 & 72.8 \;\;(7) 	\\ \hline 
$\rho$=0.40 	 & 97.8 \;\;(9)	 & 97.8 \;\;(9) 	& 78.2 (24) 	 & 72.6 \;\;(9) 	\\ \hline 
$\rho$=0.45 	 & 97.8 (11) 	 & 97.8 (11) 	 & \textbf{78.3 (16)} 	 & 72.9 \;\;(7) 	\\ \hline 
$\rho$=0.50 	 & 97.9 \;\;(8)	 & 97.9 \;\;(8) 	& 77.9 \;\;(7)	 & 72.6 \;\;(9) 	\\ \hline 
GP-TS $\rho$ 	 & 97.9 (13) 	 & 97.9 (13) 	 & 77.5 (17) 	 & 72.6 \;\;(9) 	\\ \hline 
GP-TS $\psi$ 	 & \textbf{98.0 (10)} 	 & \textbf{98.0 (10)} 	 & 77.8 (20) 	 & 72.3 \;\;(6) 	\\ \hline 
		\end{tabular}
\end{table}

These results exhibit how GP-TS pre-trains performant language models ---with top accuracy---
often at earlier interactions than when pre-training with static hyperparameters:
\eg the continually pre-trained GP-TS $\psi$ model
(see last row of Table~\ref{tab:ebay_tasks_robertabase})
provides best downstream accuracy for two e-commerce tasks and competitive accuracy in others,
in just a few pre-training interactions.

\clearpage
This efficiency is of practical importance,
due to the significant resource savings it affords.
A pre-training hyperparameter grid-search
does not provide significant downstream performance improvements,
yet it demands high computational resources
---the computational complexity of a grid-search over hyperparameters $\psi=\left(\rho, \gamma, \lambda\right)$ with $n$ candidates per hyperparameter is $\mathcal{O}(3^n)$.
On the contrary, by letting GP-TS pre-train TLMs,
best pre-training MLM performance is achieved,
with well-performing fine-tuned model accuracy across downstreams tasks,
in fewer pre-training interactions.


\section{Conclusion}
\label{sec:conclusion}
We present a multi-armed bandit-based Bayesian optimization framework for the sequential selection of pre-training hyperparameters
towards optimized Transformer-based language model performance.
We develop and evaluate an interactive, Gaussian process-based Thompson sampling (GP-TS) framework
for accelerated language model pre-training. 
We model noisy evaluations of the pre-training objective (\eg the MLM loss) as drawn from a surrogate Gaussian process that the bandit agent aims to minimize.

We provide empirical evidence of how GP-TS,
when applied to MLM dynamic masking,
attains superior and accelerated (both from-scratch and continual) pre-training performance,
along with excellent in-domain downstream metric values.
While \citet{roberta} randomly select ---with fixed probability---
which input tokens to mask,
we show that \textit{sequentially} adapting the masking hyperparameters with GP-TS results in enhanced and efficient pre-training.
Notably, GP-TS interactively selects hyperparameters that result in top performing models faster,
enabling significant resource efficiency, of critical importance in practice.

Building upon our formulation and the provided evidence,
we envision follow-up work investigating the proposed method's ability 
to successfully pre-train large-scale models in general purpose corpora,
as well as for optimizing domain-specific models.

\section*{Limitations}
%

There are several limitations to account for in the presented work.
First, the large GPU requirements for the execution and replication of the presented experiments.
Second, the lack of empirical results beyond English-based text,
and how morphologically and syntactically more complex corpora may affect the presented evidence.
Third, our evaluation section compares GP-TS performance to the common hyperparameter grid-search alternative,
yet we acknowledge that other Bayesian optimization techniques used in the machine learning community may provide suitable and competitive alternatives to explore.
In addition, we have not run any hyperparameter tuning beyond MLM dynamic masking, which might improve all studied algorithms' performance.
Finally, our conclusions are limited to RoBERTa models pre-trained via MLM dynamic masking,
and therefore, investigation of how GP-TS generalizes to other TLM pre-training approaches and architectures is lacking.

\section*{Ethics Statement}

This work raises ethical and societal considerations associated with
the use and biases of pre-collected natural language data,
the energetic and environmental impact of extensive GPU resource usage,
and the downstream applications of language models.
We acknowledge the potential implicit biases within the publicly available datasets used.
\Eg \texttt{mimic} reports are limited to the population attended at Beth Israel Deaconess Medical Center,
and may contain implicit biases of health practitioners there.
We have carefully sampled data for the \texttt{e-commerce} dataset to avoid biases over specific products, users and sellers.
We are also aware of the rising concerns pertaining to the carbon footprint of large language models~\citep{patterson2021carbon},
and the significant impact hyperparameter selection techniques have on resource utilization and power consumption~\citep{puvis-de-chavannes-etal-2021-hyperparameter}.
Finally, we acknowledge the wide range of established and anticipated risks that language models pose to society~\citep{j-Weidinger2021}.

\section*{Acknowledgements}
I\~nigo Urteaga and Moulay-Za\"idane Dra\"idia were partially supported
by funds from eBay's Research and University Partnership for Technology (eRUPT) program.
We also acknowledge computing resources from Columbia University’s Shared Research Computing Facility project,
which is supported by NIH Research Facility Improvement Grant 1G20RR030893-01,
and associated funds from the New York State Empire State Development,
Division of Science Technology and Innovation (NYSTAR) Contract C090171.
both awarded April 15, 2010.


\clearpage
\appendix
\onecolumn
\appendix

\clearpage
\section{Appendix: Gaussian process details}
\label{asec:GP_details}

\paragraph*{Gaussian processes.}\hspace*{-2ex}
A GP is a stochastic process, ${f(\psi) : \psi \in \Psi }$, such that
for any finite set of elements $\psi_1, \cdots , \psi_k \in \Psi$,
the associated finite collection of random variables $f(\psi_1), \cdots, f(\psi_k)$, 
has a multivariate Gaussian distribution~\citep{b-Rasmussen2005}.

A GP $f(\psi) \sim GP(\mu(\cdot), k(\cdot,\cdot))$ can be understood as a probability distribution over arbitrary functions, with $\mu(\psi) = \mathbb{E}[f(\psi)]$ its mean function, and $k(\cdot, \cdot)$ the covariance kernel, \ie $k(\psi, \psi^\prime)=\mathbb{E}[(f(\psi)-\mu(\psi))^\top(f(\psi^\prime)-\mu(\psi^\prime))]$.

The mean and kernel functions determine the GP function class: \ie the regularity and smoothness assumptions of the modeled data.
These are parameterized prior-functions $\mu(\cdot|\theta_{\mu})$ and $k(\cdot, \cdot|\theta_k)$,
which can be fitted to the observed data $r_{1:T} = (r_1, \cdots, r_T)$ at inputs $\psi_{1:T} = (\psi_1, \cdots, \psi_T)$.

For instance, via Type-II maximum likelihood estimation (MLE) of the GP model's hyperparameters $\theta=(\theta_{\mu}, \theta_k)$,
$
\hat{\theta} =\argmax_{\theta} \log p\left(r_{1:T}|f(\psi_{1:T} | \theta) \right) 
$,
where the data likelihood $p(r|f (\cdot; \theta))$ is a function of the observation noise's probability distribution.
Bayesian approaches to hyperparameter selection for GP model training can also be implemented~\citep{b-Rasmussen2005}.

\paragraph*{Gaussian process posteriors.}\hspace*{-2ex}
Given a fitted GP, posterior inference
---computing the predictive distribution of a new datapoint $\psi^\prime$ after observing $\psi_{1:T}$---
can be performed in closed form for the Gaussian observation noise case.
For example, when the noise in Equation~\eqref{eq:rewards_gp} is \iid drawn from $\epsilon_t \sim \N{\epsilon | 0, \sigma_{\epsilon}^2}$.

Formally, given a set of observations $r_{1:T}$ at inputs $\psi_{1:T}$,
the posterior distribution of $f$ is a GP with the following mean and covariance functions:
\begin{align}
&\mu_T(\psi) = k_T(\psi)^\top (K_T + \sigma_{\epsilon}^2 I)^{-1}r_{1:T} \; , \nonumber \\
&k_T(\psi, \psi^\prime) = k(\psi,\psi^\prime) - k_T(\psi)^\top (K_T + \sigma_{\epsilon}^2 I)^{-1} k_T(\psi^\prime) \;, \nonumber \\
&\text{with}
\begin{cases}
k_T (\psi) = \left( k(\psi_1, \psi), \cdots, k(\psi_T, \psi)\right)^\top \;,\\
K_T = \left( k(\psi,\psi^\prime) \right)_{\forall \psi, \psi^\prime \in \psi_{1:T}} \;.
\end{cases} 
\label{eq:gp_posterior}
\end{align}
These closed-form posterior inference expressions can be efficiently computed, both in exact and approximate ways~\citep{b-Rasmussen2005,ip-Pleiss2018}.
Posterior inference with observation noise beyond the Gaussian assumption is an active research area, with many approximate techniques available for practitioners~\citep{ic-Snelson2006,ip-Titsias2009,ip-Wilson2015,ip-Flaxman2015}.

\clearpage
\section{Appendix: Implementation and experimentation details}
\label{asec:implementation_details}

\subsection{Gaussian process}
\label{asec:implementation_details_gp}

We implement Gaussian process modules based on GPyTorch~\citep{gpytorch},
and execute all experiments with a GP process prior and GP fitting details as described in Table~\ref{tab:gp_prior}.
\begin{table}[!h]
	\caption{Gaussian Process prior and hyperparameters.}
	\vspace*{-2ex}
	\label{tab:gp_prior}
	\begin{center}
		\begin{tabular}{|c|c|}
			\hline
			Hyperparameter\cellcolor[gray]{0.6} & Initial Value \cellcolor[gray]{0.6} \\ \hline
\multicolumn{2}{|c|}{\cellcolor[gray]{0.9} GP Model}\\ \hline 
Mean Function & Constant \\ \hline 
Prior constant & 0 \\ \hline 
Kernel Function & Scaled RBF Kernel \\ \hline 
Prior output-scale & 1 \\ \hline 
Prior length-scale & 0.25 \\ \hline
\multicolumn{2}{|c|}{\cellcolor[gray]{0.9} Observation Model}\\ \hline 
Likelihood function & Gaussian \\ \hline 
Noise variance & 1 \\ \hline 
\multicolumn{2}{|c|}{\cellcolor[gray]{0.9} Training details}\\ \hline 
Loss function & ExactMarginalLogLikelihood \\ \hline 
train max iters & 100 \\ \hline 
loss epsilon & 0.01 \\ \hline 
\multicolumn{2}{|c|}{\cellcolor[gray]{0.9} Optimizer } \\ \hline 
optimizer & adam \\ \hline 
lr & 0.1 \\ \hline 
		\end{tabular}
	\end{center}
\end{table}

We take the most conservative approach on GP-TS prior and hyperparameter selection:
we utilize an uninformative prior, with no preference for any hyperparameter configuration.
This is the less assuming yet more challenging experimental set-up,
where we evaluate whether GP-TS can successfully learn ---without any prior knowledge--- to find good hyperparameters.

Based on bandit theory and practice,
informative priors can accelerate convergence if properly specified
(\ie when more mass is put into favorable regions of the hyperparameter space);
while slowing down convergence, if incorrectly specified
(\ie when mass is put in unfavorable regions of the space).
Evaluating how different priors affect GP-TS are experiments left as future work.

\clearpage
\subsection{RoBERTa pre-training}
\label{asec:implementation_details_roberta_pretrain}

We pre-train all RoBERTa models as provided by~\citet{fairseq},
with the BERT-base architecture of 125M parameters, by minimizing the MLM loss with dynamic masking
in a server with 8 Tesla V100-SXM2-32GB GPUs.
We execute the RoBERTa pre-training procedure as described in Fairseq's RoBERTa pre-training tutorial\footnote{
	Available at \url{https://github.com/pytorch/fairseq/blob/main/examples/roberta/README.pretraining.md}
},
with specific hyperparameters as described in Table~\ref{tab:roberta_pretrain}.

The interactions for \texttt{wiki-c4} and \texttt{e-commerce} contain 1000 updates each (\ie $u=1000$), while we reduce the number of updates per-interaction to $u=500$ when pre-training with \texttt{mimic} notes.

\begin{table}[!h]
	\caption{RoBERTa pre-training hyperparameters.}
	\vspace*{-2ex}
	\label{tab:roberta_pretrain}
	\begin{center}
		\begin{tabular}{|c|c|}
			\hline
			Hyperparameter\cellcolor[gray]{0.6} & Value \cellcolor[gray]{0.6} \\ \hline
Architecture & RoBERTa base \\ \hline 
Task& masked lm \\ \hline 
Criterion & masked lm \\ \hline 
\multicolumn{2}{|c|}{\cellcolor[gray]{0.9} Model details}\\ \hline 
dropout & 0.1 \\ \hline 
attention-dropout & 0.1 \\ \hline 
weight-decay & 0.01 \\ \hline 
\multicolumn{2}{|c|}{\cellcolor[gray]{0.9} Training details}\\ \hline 
batch-size & 32 \\ \hline 
update-freq & 16 \\ \hline 
sample-break-mode & complete \\ \hline 
tokens-per-sample & 512 \\ \hline  
\multicolumn{2}{|c|}{\cellcolor[gray]{0.9} Optimizer } \\ \hline 
optimizer &adam \\ \hline 
adam-betas & (0.9,0.98) \\ \hline 
adam-eps & 1e-6 \\ \hline 
clip-norm & 1.0 \\ \hline 
\multicolumn{2}{|c|}{\cellcolor[gray]{0.9} Learning rate} \\ \hline 
lr &0.0005 \\ \hline 
lr-scheduler & polynomial decay \\ \hline 
linear-warmup-updates & 1000 \\ \hline 
\multicolumn{2}{|c|}{\cellcolor[gray]{0.9} Dynamic masking } \\ \hline 
mask-prob & $\rho$ \\ \hline 
leave-unmasked-prob & 0.1 \\ \hline 
random-token-prob & 0.1 \\ \hline 
		\end{tabular}
	\end{center}
\end{table}

\clearpage
\subsection{Summary statistics of the computational cost}
\label{asec:computational_overhead}

We provide in Table~\ref{tab:pretraining_compcost} summary statistics
on the execution time of GP-TS pre-training in our experiments,
as per details in Section~\ref{asec:implementation_details_roberta_pretrain}.
The per-interaction, average execution time of pre-training is:
33,316 seconds for the \texttt{wiki-c4} dataset;
37,392 seconds for the \texttt{e-commerce} data;
and 1,489 seconds for \texttt{MIMIC} notes.
It only takes about 20 seconds on average to execute GP-TS per-interaction.
Hence, the overhead is of 0.05\% for the biggest dataset, and 1\% for the smallest one.
We note that the TLM pre-training implementation of~\citet{fairseq} leverages GPU computations,
while GP-TS is executed within a single CPU ---with no GPU acceleration.


\begin{table}[!h]
	\caption{Per-interaction execution time of TLM pre-training and GP-TS: average time in seconds, plus-minus the standard deviation.}
	\vspace*{-2ex}
	\label{tab:pretraining_compcost} 
	\begin{center}
		\begin{tabular}{|c|c|c|c|}\hline
			 \cellcolor[gray]{0.5} &  \multicolumn{3}{c|}{Execution time in seconds \cellcolor[gray]{0.5} } \\ \cline{2-4}
			\multirow{-2}{*}{\cellcolor[gray]{0.5}Dataset} & TLM Pre-training \cellcolor[gray]{0.6} & GP-TS $\rho$ \cellcolor[gray]{0.6} & GP-TS $\psi$ \cellcolor[gray]{0.6} \\ \hline
			\cellcolor[gray]{0.95} \texttt{wiki-c4} 
				& $33,316 \pm 395 \; s$ \cellcolor[gray]{0.95} &  $19 \pm 6 \; s$  \cellcolor[gray]{0.95} &  $21 \pm 6 \; s$ \cellcolor[gray]{0.95} \\  \hline 
			\cellcolor[gray]{1.0} \texttt{mimic} 
				& $1489 \pm 46 \; s$ \cellcolor[gray]{1.0} &  $16 \pm 5 \; s$ \cellcolor[gray]{1.0} & $17 \pm 5 \; s$ \cellcolor[gray]{1.0} \\  \hline 
			\cellcolor[gray]{0.95} \texttt{e-commerce}
				& $37,392 \pm 494 \; s$ \cellcolor[gray]{0.95} & $21 \pm 3 \; s$ \cellcolor[gray]{0.95} & $23 \pm 10 \; s$ \cellcolor[gray]{0.95} \\  \hline 
		\end{tabular}
	\end{center}
\end{table}

\subsection{Summary statistics of the pre-training datasets}
\label{asec:pretraining_dataset_details}

We split each pre-training dataset into 80\%-10\%-10\% training, validation and test sets for our experiments, with summary statistics of each set provided in Table~\ref{tab:pretraining_dataset_details}.


\begin{table}[!h]
	\caption{Summary statistics of the pre-training datasets.}
	\vspace*{-2ex}
	\label{tab:pretraining_dataset_details} 
	\begin{center}
		\begin{tabular}{|c|c|c|c|}\hline
			\multicolumn{2}{c|}{Dataset \cellcolor[gray]{0.6}} & Total word count \cellcolor[gray]{0.6}  & Average words per sentence \cellcolor[gray]{0.6} \\ \hline
			\cellcolor[gray]{0.95} & Training \cellcolor[gray]{0.95} &  4,517,625,794 \cellcolor[gray]{0.95} &  \; 35.9 \cellcolor[gray]{0.95} \\  
			\cellcolor[gray]{0.95} & Validation \cellcolor[gray]{0.95} & \; 735,950,955\cellcolor[gray]{0.95} & \; 35.6 \cellcolor[gray]{0.95} \\  
			\multirow{-3}{*}{\cellcolor[gray]{0.95} \texttt{wiki-c4} } & Test \cellcolor[gray]{0.95} & \; 735,571,833 \cellcolor[gray]{0.95}& \; 35.6 \cellcolor[gray]{0.95} \\ \hline 
			\cellcolor[gray]{1.0} & Training \cellcolor[gray]{1.0} &  \; 402,720,632 \cellcolor[gray]{1.0} & 216.7 \cellcolor[gray]{1.0} \\  
			\cellcolor[gray]{1.0} & Validation \cellcolor[gray]{1.0} & \;\;\;  82,340,235 \cellcolor[gray]{1.0} & 658.7 \cellcolor[gray]{1.0} \\  
			\multirow{-3}{*}{\cellcolor[gray]{1.0} \texttt{mimic} } & Test \cellcolor[gray]{1.0} & \;\;  18,735,884  \cellcolor[gray]{1.0}& 187.3 \cellcolor[gray]{1.0} \\ \hline 
			\cellcolor[gray]{0.95} & Training \cellcolor[gray]{0.95} & 3,935,845,017 \cellcolor[gray]{0.95} & \;\; 5.6 \cellcolor[gray]{0.95} \\  
			\cellcolor[gray]{0.95} & Validation \cellcolor[gray]{0.95} & \;\;  494,802,278 \cellcolor[gray]{0.95} & \;\; 5.5 \cellcolor[gray]{0.95} \\  
			\multirow{-3}{*}{\cellcolor[gray]{0.95} \texttt{e-commerce} } & Test \cellcolor[gray]{0.95} & \;\;  482,733,197 \cellcolor[gray]{0.95} & \;\; 5.5 \cellcolor[gray]{0.95} \\ \hline 
		\end{tabular}
	\end{center}
\end{table}

\newpage
\subsection{RoBERTa fine-tuning}
\label{asec:implementation_details_roberta_fine-tune}

The specific RoBERTa hyperparameters used for the in-domain fine-tuning downstream tasks are described in Tables~\ref{tab:roberta_finetune_eclassification}--\ref{tab:roberta_finetune_medical}.

\begin{table}[!h]
	\caption{RoBERTa fine-tuning hyperparameters for the e-commerce title classification downstream task.}
	\label{tab:roberta_finetune_eclassification}
	\begin{center}
		\begin{tabular}{|c|c|}
			\hline
			Hyperparameter\cellcolor[gray]{0.6} & Value \cellcolor[gray]{0.6} \\ \hline
Architecture & RoBERTa base \\ \hline 
\multicolumn{2}{|c|}{\cellcolor[gray]{0.9} Task}\\ \hline 
Task& sentence prediction \\ \hline 
Criterion & sentence prediction \\ \hline 
num-classes & 2 \\ \hline 
max-positions & 512 \\ \hline
init-token & 0  \\ \hline 
separator-token & 2  \\ \hline 
\multicolumn{2}{|c|}{\cellcolor[gray]{0.9} Model details}\\ \hline 
dropout & 0.1 \\ \hline 
attention-dropout & 0.1 \\ \hline 
\multicolumn{2}{|c|}{\cellcolor[gray]{0.9} Dataset}\\ \hline 
batch-size & 32 \\ \hline 
update-freq & 1 \\ \hline 
required-batch-size-multiple & 1 \\ \hline
max-tokens & 4400 \\ \hline
skip-invalid-size-inputs-valid-test & True \\ \hline
\multicolumn{2}{|c|}{\cellcolor[gray]{0.9} Optimizer } \\ \hline 
optimizer &adam \\ \hline 
weight-decay & 0.1 \\ \hline 
adam-betas & (0.9,0.98) \\ \hline 
adam-eps & 1e-6 \\ \hline 
\multicolumn{2}{|c|}{\cellcolor[gray]{0.9} Learning rate} \\ \hline 
lr-scheduler & polynomial decay \\ \hline 
lr & 1e-5 \\ \hline 
linear-warmup-updates & 1000 \\ \hline 
max-updates & 100000 \\ \hline 
max-epoch & 10 \\ \hline
clip-norm & 0.0 \\ \hline 
		\end{tabular}
	\end{center}
\end{table}
\begin{table}[!h]
	\caption{RoBERTa fine-tuning hyperparameters for the e-commerce title similarity downstream task.}
	\label{tab:roberta_finetune_esimilarity}
	\begin{center}
		\begin{tabular}{|c|c|}
			\hline
			Hyperparameter\cellcolor[gray]{0.6} & Value \cellcolor[gray]{0.6} \\ \hline
Architecture & RoBERTa base \\ \hline 
\multicolumn{2}{|c|}{\cellcolor[gray]{0.9} Task}\\ \hline 
Task& sentence prediction \\ \hline 
Criterion & sentence prediction \\ \hline 
num-classes & 2 \\ \hline 
max-positions & 512 \\ \hline
init-token & 0  \\ \hline 
separator-token & 2  \\ \hline 
\multicolumn{2}{|c|}{\cellcolor[gray]{0.9} Model details}\\ \hline 
dropout & 0.1 \\ \hline 
attention-dropout & 0.1 \\ \hline 
\multicolumn{2}{|c|}{\cellcolor[gray]{0.9} Dataset}\\ \hline 
batch-size & 32 \\ \hline 
update-freq & 1 \\ \hline 
required-batch-size-multiple & 1 \\ \hline
max-tokens & 4400 \\ \hline
skip-invalid-size-inputs-valid-test & True \\ \hline
\multicolumn{2}{|c|}{\cellcolor[gray]{0.9} Optimizer } \\ \hline 
optimizer &adam \\ \hline 
weight-decay & 0.1 \\ \hline 
adam-betas & (0.9,0.98) \\ \hline 
adam-eps & 1e-6 \\ \hline 
\multicolumn{2}{|c|}{\cellcolor[gray]{0.9} Learning rate} \\ \hline 
lr-scheduler & polynomial decay \\ \hline 
lr & 1e-5 \\ \hline 
linear-warmup-updates & 1000 \\ \hline 
max-updates & 100000 \\ \hline 
max-epoch & 10 \\ \hline
clip-norm & 0.0 \\ \hline 
		\end{tabular}
	\end{center}
\end{table}
\begin{table}[!h]
	\caption{RoBERTa fine-tuning hyperparameters for the e-commerce title quality downstream task.}
	\label{tab:roberta_finetune_equality}
	\begin{center}
		\begin{tabular}{|c|c|}
			\hline
			Hyperparameter\cellcolor[gray]{0.6} & Value \cellcolor[gray]{0.6} \\ \hline
Architecture & RoBERTa base \\ \hline 
\multicolumn{2}{|c|}{\cellcolor[gray]{0.9} Task}\\ \hline 
Task& sentence prediction \\ \hline 
Criterion & sentence prediction \\ \hline 
num-classes & 2 \\ \hline 
max-positions & 512 \\ \hline
init-token & 0  \\ \hline 
separator-token & 2  \\ \hline 
\multicolumn{2}{|c|}{\cellcolor[gray]{0.9} Model details}\\ \hline 
dropout & 0.1 \\ \hline 
attention-dropout & 0.1 \\ \hline 
\multicolumn{2}{|c|}{\cellcolor[gray]{0.9} Dataset}\\ \hline 
batch-size & 32 \\ \hline 
update-freq & 1 \\ \hline 
required-batch-size-multiple & 1 \\ \hline
max-tokens & 4400 \\ \hline
skip-invalid-size-inputs-valid-test & True \\ \hline
\multicolumn{2}{|c|}{\cellcolor[gray]{0.9} Optimizer } \\ \hline 
optimizer &adam \\ \hline 
weight-decay & 0.1 \\ \hline 
adam-betas & (0.9,0.98) \\ \hline 
adam-eps & 1e-6 \\ \hline 
\multicolumn{2}{|c|}{\cellcolor[gray]{0.9} Learning rate} \\ \hline 
lr-scheduler & polynomial decay \\ \hline 
lr & 1e-5 \\ \hline 
linear-warmup-updates & 1000 \\ \hline 
max-updates & 100000 \\ \hline 
max-epoch & 10 \\ \hline
clip-norm & 0.0 \\ \hline 
		\end{tabular}
	\end{center}
\end{table}
\begin{table}[!h]
	\caption{RoBERTa fine-tuning hyperparameters for the medical MLI downstream task.}
	\label{tab:roberta_finetune_medical}
	\begin{center}
		\begin{tabular}{|c|c|}
			\hline
			Hyperparameter\cellcolor[gray]{0.6} & Value \cellcolor[gray]{0.6} \\ \hline
Architecture & RoBERTa base \\ \hline 
\multicolumn{2}{|c|}{\cellcolor[gray]{0.9} Task}\\ \hline 
Task& sentence prediction \\ \hline 
Criterion & sentence prediction \\ \hline 
num-classes & 3 \\ \hline 
max-positions & 512 \\ \hline
init-token & 0  \\ \hline 
separator-token & 2  \\ \hline 
\multicolumn{2}{|c|}{\cellcolor[gray]{0.9} Model details}\\ \hline 
dropout & 0.1 \\ \hline 
attention-dropout & 0.1 \\ \hline 
\multicolumn{2}{|c|}{\cellcolor[gray]{0.9} Dataset}\\ \hline 
batch-size & 32 \\ \hline 
update-freq & 1 \\ \hline 
required-batch-size-multiple & 1 \\ \hline
max-tokens & 4400 \\ \hline
skip-invalid-size-inputs-valid-test & True \\ \hline
\multicolumn{2}{|c|}{\cellcolor[gray]{0.9} Optimizer } \\ \hline 
optimizer &adam \\ \hline 
weight-decay & 0.1 \\ \hline 
adam-betas & (0.9,0.98) \\ \hline 
adam-eps & 1e-6 \\ \hline 
\multicolumn{2}{|c|}{\cellcolor[gray]{0.9} Learning rate} \\ \hline 
lr-scheduler & polynomial decay \\ \hline 
lr & 1e-5 \\ \hline 
linear-warmup-updates & 1000 \\ \hline 
max-updates & 100000 \\ \hline 
max-epoch & 10 \\ \hline
clip-norm & 0.0 \\ \hline 
		\end{tabular}
	\end{center}
\end{table}

\clearpage
\subsection{Summary statistics of the fine-tuning datasets}
\label{asec:finetuning_dataset_details}

We split each per-task fine-tuning dataset into training, development and test sets for our experiments, with summary statistics of each set provided in Table~\ref{tab:finetuning_dataset_details}.


\begin{table}[!h]
	\caption{Summary statistics of the fine-tuning task datasets.}
	\vspace*{-2ex}
	\label{tab:finetuning_dataset_details} 
	\begin{center}
		\begin{tabular}{|c|c|c|c|}\hline
			\multicolumn{2}{c|}{Dataset \cellcolor[gray]{0.6}} & Total sentence count \cellcolor[gray]{0.6}  & \shortstack{Average words per sentence \\ Input0 -- Input1} \cellcolor[gray]{0.6} \\ \hline
			\cellcolor[gray]{0.95} & Training  \cellcolor[gray]{0.95} & 224,745 \cellcolor[gray]{0.95} & 10.9 -- 10.9\cellcolor[gray]{0.95} \\  
			\cellcolor[gray]{0.95} & Dev \cellcolor[gray]{0.95} & \;\; 6,035 \cellcolor[gray]{0.95} & 10.9 -- 10.8 \cellcolor[gray]{0.95} \\  
			\multirow{-3}{*}{\cellcolor[gray]{0.95} \shortstack{\texttt{e-commerce title} \\ \texttt{classification \& similarity}} } & Test \cellcolor[gray]{0.95} &  12,311 \cellcolor[gray]{0.95}& 10.9 -- 10.8 \cellcolor[gray]{0.95} \\ \hline 
			\cellcolor[gray]{1.0} & Training \cellcolor[gray]{1.0} & 49,420 \cellcolor[gray]{1.0} & 10.6 -- NA \cellcolor[gray]{1.0} \\  
			\cellcolor[gray]{1.0} & Dev \cellcolor[gray]{1.0} & \; 2,629 \cellcolor[gray]{1.0} & 9.8 -- NA \cellcolor[gray]{1.0} \\  
			\multirow{-3}{*}{\cellcolor[gray]{1.0} \texttt{e-commerce title quality} } & Test \cellcolor[gray]{1.0} & 5,174 \cellcolor[gray]{1.0}& 9.8 -- NA \cellcolor[gray]{1.0} \\ \hline 
			\cellcolor[gray]{0.95} & Training \cellcolor[gray]{0.95} & 11,232 \cellcolor[gray]{0.95} & 15.9 -- 5.5\cellcolor[gray]{0.95} \\  
			\cellcolor[gray]{0.95} & Dev \cellcolor[gray]{0.95} & \; 1,395 \cellcolor[gray]{0.95} & 16.9 -- 5.4 \cellcolor[gray]{0.95} \\  
			\multirow{-3}{*}{\cellcolor[gray]{0.95} \texttt{medical MLI} } & Test \cellcolor[gray]{0.95} & \; 1,422 \cellcolor[gray]{0.95} & 15.4 -- 5.4 \cellcolor[gray]{0.95} \\ \hline 
		\end{tabular}
	\end{center}
\end{table}


\end{document}